\documentclass[10pt,journal]{IEEEtran}
\usepackage{amsmath,amsfonts}
\usepackage{algorithmic}
\usepackage{algorithm}
\usepackage{array}
\usepackage{textcomp}
\usepackage{stfloats}
\usepackage{url}
\usepackage{verbatim}
\usepackage{graphicx}
\usepackage{subcaption}
\usepackage{cases,multirow}
\usepackage[numbers,sort&compress]{natbib}
\usepackage{booktabs}
\usepackage{tabularx}
\usepackage[colorlinks=true,linkcolor=blue,citecolor=blue]{hyperref}
\usepackage{bbm}
\usepackage{color}

\usepackage{tikz}
\usetikzlibrary{shapes.geometric, arrows, positioning, fit, calc}

\graphicspath{{figures/}}

\newcommand{\bE}{\mathbb E}
\newcommand{\bS}{\mathbb S}
\newcommand{\bR}{\mathbb R}

\newcommand{\bM}{\mathbb M}

\newcommand{\tgamma}{\Gamma}

\newcommand{\cU}{\mathcal U}

\newcommand{\cC}{\mathcal C}
\newcommand{\cF}{\mathcal F}

\newcommand{\dtheta}{\dot\theta}

\newcommand{\fp}{\mathbf p}
\newcommand{\fx}{\mathbf x}

\newcommand{\fs}{\mathbf s}

\newcommand{\dfx}{\dot{\mathbf{x}}}
\newcommand{\dx}{\dot x}

\begin{document}

\title{Dynamic Exploration on Segment-Proposal Graphs for Tubular Centerline Tracking}

%\author{IEEE Publication Technology,~\IEEEmembership{Staff,~IEEE,}
\author{Chong Di, Jinglin Zhang, Zhenjiang Li, Jean-Marie Mirebeau, Da Chen, Laurent D. Cohen,~\IEEEmembership{Fellow,~IEEE}
        % <-this % stops a space
 \IEEEcompsocitemizethanks{
\IEEEcompsocthanksitem Chong Di is  with Shandong Artificial Intelligence Institute, Qilu University of Technology (Shandong Academy of Sciences), Jinan, China. e-mail: cdi@qlu.edu.cn\protect %\\
\IEEEcompsocthanksitem Jinglin Zhang is  with School of Control Science and Engineering, Shandong University, Jinan, China. e-mail: jinglin.zhang@sdu.edu.cn\protect %\\
\IEEEcompsocthanksitem Jean-Marie Mirebeau is with Department of Mathematics, Centre Borelli, ENS Paris-Saclay, CNRS, University Paris-Saclay, 91190, Gif-sur-Yvette, France. e-mail: jean-marie.mirebeau@ens-paris-saclay.fr\protect%\\
\IEEEcompsocthanksitem Zhenjiang~Li is with Department of Radiation Oncology, Shandong Cancer Hospital and Institute, Shandong First Medical University, Shandong Academy of  Medical Sciences, 250117 Jinan, China. (e-mail:zhenjli1987@163.com)\protect
\IEEEcompsocthanksitem Da Chen and Laurent D. Cohen are with CEREMADE, University Paris Dauphine, Universit\'e-PSL, CNRS, UMR 7534, 75775 Paris, France. e-mail: chenda@ceremade.dauphine.fr, cohen@ceremade.dauphine.fr\protect\\
%Corresponding author: Da Chen.
}% <-this % stops an unwanted space       
\thanks{}% <-this % stops a space
%\thanks{Manuscript received April 19, 2021; revised August 16, 2021.}
}



%\maketitle
\IEEEtitleabstractindextext{
\begin{abstract}
Optimal curve methods provide a fundamental framework for tubular centerline tracking. Point-wise approaches, such as minimal paths, are theoretically elegant but often suffer from shortcut and short-branch combination problems in complex scenarios. Nonlocal segment-wise methods address these issues by mapping pre-extracted centerline fragments onto a segment-proposal graph, performing optimization in this abstract space, and recovering the target tubular centerline from the resulting optimal path. In this paradigm, graph construction is critical, as it directly determines the quality of the final result. However, existing segment-wise methods construct graphs in a static manner, requiring all edges and their weights to be pre-computed i.e. the graph must be sufficiently complete prior to search. Otherwise, the true path may be absent from the candidate space, leading to search failure. To address this limitation, we propose a dynamic exploration scheme for constructing segment-proposal graphs, where the graph is built on demand during the search for optimal paths. By formulating the problem as a Markov decision process, we apply Q-learning to compute edge weights only for visited transitions and adaptively expand the action space when connectivity is insufficient. Experimental results on retinal vessels, roads, and rivers demonstrate consistent improvements over state-of-the-art methods in both accuracy and efficiency.
%The computation of minimal paths for the applications in tracking tubular structures such as blood vessels and roads is challenged by complex morphologies and environmental variations. Existing approaches can be roughly categorized into  two research lines: the point-wise based models and the segment-wise based models. Although segment-wise approaches have obtained promising results in many scenarios, they often suffer from computational inefficiency and rely heavily on a prescribed prior to fit the target elongated shapes. We propose a novel framework that casts segment-wise tracking as a Markov decision process, enabling a reinforcement learning approach. Our method leverages Q-Learning to dynamically explore a graph of segments, computing edge weights on-demand and adaptively expanding the search space. This strategy avoids the high cost of a pre-computed graph and proves robust to incomplete initial information. Experimental results on typical tubular structure datasets demonstrate that our method significantly outperforms state-of-the-art point-wise and segment-wise approaches. The proposed method effectively handles complex topologies and maintains global path coherence without depending on extensive prior structural knowledge.
\end{abstract}

}

\maketitle
\IEEEdisplaynontitleabstractindextext

\IEEEpeerreviewmaketitle

\section{Introduction}\label{sec:introduction}

\begin{figure*}[t]
\centering
\includegraphics[width=\textwidth]{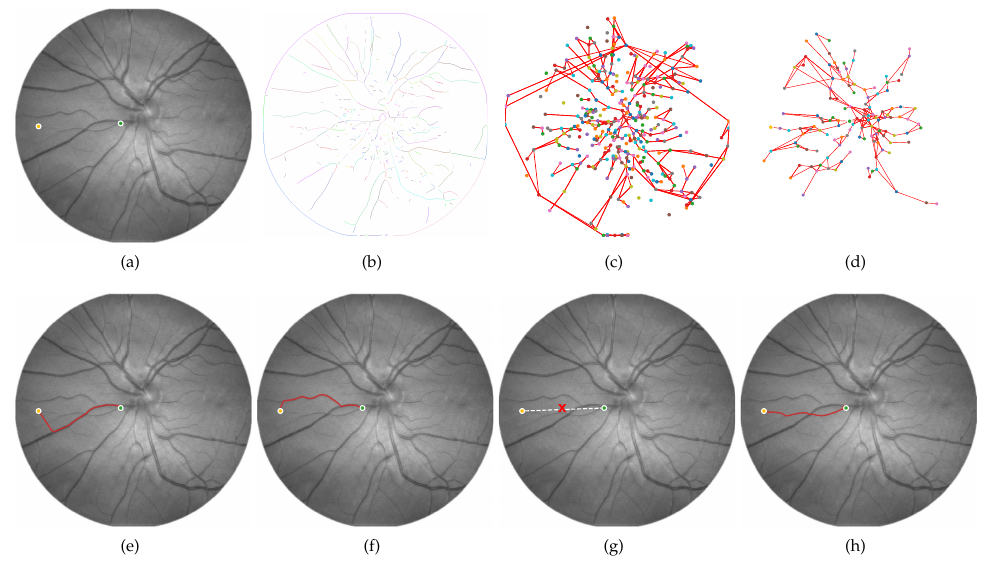}
\caption{Demonstration of the proposed method on a retinal image and comparison with other approaches. (\textbf{a}) Input image with start (green) and end (yellow) points. (\textbf{b}) Generated disjoint centerline segments. (\textbf{c}) The initial graph that traditional segment-wise methods must construct, requiring all edge weights to be computed. (\textbf{d}) Our method dynamically explores a much smaller subgraph, saving significant computational cost. (\textbf{e}) and (\textbf{f}) Point-wise methods (anisotropic and curvature-penalized, respectively) struggle to navigate through high-contrast regions. (\textbf{g}) Existing segment-wise methods fail when the initial graph is incomplete and does not contain a path from source to destination. (\textbf{h}) Our proposed method successfully finds the correct minimal path by adaptively exploring the graph.}
\label{fig:example_minimal_path}
\end{figure*}

Tubular structures such as blood vessels in medical images and roads in aerial images, constitute a prevalent topological pattern in both natural and artificial environments~\cite{li2022robust,qi2021examinee,song2025optimized,moccia2018blood,liu2023energy}. In general, interactive  tubular tracing based on optimal curves aims to seek continuous paths  connecting two landmark points~\cite{cohen1997global,peyre2010geodesic,mirebeau2019hamiltonian} as tubular centerlines. This capability is central to applications including vascular pathway visualization for medical diagnosis, road-network planning, and hydrological navigation~\cite{rossant2024characterization,pore2023autonomous,jiang2021autonomous}. The computational difficulty is compounded by heterogeneous tubular morphologies and highly variable surrounding contexts.

Basically, a great variety of tubular structure tracking approaches are established in a manner of sequential accumulation of point-wise tubular features from a given source point, in conjunction with a geometric regularization term. Cohen and Kimmel introduce a foundational model~\cite{cohen1997global} for computing minimal paths based on the Eikonal equation framework embedded with a Riemannian metric, allowing to apply the efficient fast marching method~\cite{sethian1996fast,sethian1999fast,mirebeau2019riemannian} as the numerical solver. This original minimal model has inspired a series of applications for tracking tubular structures~\cite{moriconi2018inference,bertrand2024fitting,jain2021morphological,kaul2012detecting,turetken2016reconstructing}.  However,  the geodesic metrics considered in this work are not particularly designed for tracking minimal paths passing through tubular structure centerlines. A subsequent approach led to two-stage vessel centerline tracking model which is able to extract centerline information from a prescribed tubularity segmentation procedure~\cite{deschamps2001fast}. An alternative idea for addressing the issue in~\cite{cohen1997global} is to model a tubular structure via its centerline and the radius at each centerline point. This yields two elegant minimal path-based vessel tracking models~\cite{benmansour2011tubular,li2007vessels}, which integrate local radius as an additional dimension. Furthermore, the model~\cite{benmansour2011tubular} also takes into account the orientations of tubular structures as geometric enhancement for anisotropic Riemannian metrics. 

Planar path curvature is known as an important and useful  property for characterizing the geometry of tubular structure centerlines. Minimal path approaches~\cite{chen2017global,mirebeau2018fast,duits2018optimal,bekkers2015pde,chen2023computing,van2024geodesic},  which exploit curvature-based penalty as model regularization and are founded over an orientation-lifted space,  are capable of computing optimal paths featuring rigid and elastic properties. Compared to the first-order models mentioned above, those curvature-penalized models can reduce the possibility of shortcuts and short branches combination problems. In addition, the GPU implementation~\cite{mirebeau2023massively} can greatly reduce the computation time for computing curvature-penalized minimal paths, leading to  practical applications in tracking tubular structure centerlines. Liao~\emph{et al.}~\cite{liao2018progressive} introduced a path selective model based on a front freezing scheme~\cite{cohen2007segmentation} in terms of features which are estimated from short back-tracked paths. During the computation of the geodesic distances, a set of points might be tagged as obstacles so as to prevent the distance front pass through unexpected locations. In~\cite{chen2019minimal}, the authors propose to simultaneously update the geodesic distances and the metrics, where the evolution of the geodesic metrics is implemented via a non-local geometric features from the image data. As a result, such a dynamic Riemannian metric-based minimal path model allows to take the tubular appearance coherence prior into consideration. More recently, nonlocal pattern information has been integrated with local curvature regularization to enhance the detection of attenuated structures in complex environments~\cite{liu2023curvilinear}. Despite their theoretical elegance, the aforementioned minimal path models predicated on the Hamiltonian-Jacobi-Bellman (HJB) partial differential equation (PDE) framework exhibit significant limitations. For example, the short branch combination problem still persists when confronting complex morphological configurations. 

Beyond methods that rely solely on geodesic paths, deep convolutional neural networks have also been explored as an efficient tool to provide semantic-level information, showing promising results  for tracking tubular structure centerlines. For instance, the Path-CNN model~\cite{liao2022progressive} was introduced to integrate convolutional neural networks as a binary classifier into the HJB PDE-based minimal path computation process. During the geodesic distance propagation, a part of each geodesic path emanating from the front is recognized via the classifier, such that a front point will be frozen (i.e. set as a obstacle point) if the corresponding back-tracked path is unsatisfactory. In order to improve the performance of the classifier, an iterative training protocol with specially crafted samples was developed to concurrently extract centerlines and segmentation masks while mitigating discrepancies between training and inference samples~\cite{liao2023segmentation}. 

Nonlocal segment-wise approaches consider tubular centerlines as combinations of piecewise curvilinear segments extracted from a set of centerline proposals. Specifically,  Saumya \emph{et al.}~\cite{saumya2023Topology} proposed to detect a set of connected piecewise tubular structure centerline segments using the morse theorem, each of which is assigned to a value indicating the probability of this segment belongs to a tubular structure. This method avoids to predict the probability for each point in the image domain and has shown promising results.  Nevertheless, these neural network-based approaches necessitate extensive training datasets and frequently encounter limitations in generalizability across diverse imaging contexts. 

In contrast to point-wise minimal path tracking methods, segment-wise techniques mitigate the shortcut problem by leveraging pre-extracted segments that inherently capture local geometry and structural coherence. An interactive approach for retinal vessel extraction, for example, integrates vessel tracing with graph search, constructing graphs from curve fragments to overcome limitations in extracting tortuous vessels and reducing interference between neighboring vessels~\cite{wang2013interactive}. This method reduces the need for user intervention in low-contrast images while enhancing computational efficiency. Another approach introduced a trajectory grouping method with curvature regularization, which combines prescribed tubular segments with curvature-penalized geodesic paths~\cite{liu2021trajectory}. By integrating local smoothness priors with global optimization through graph-based path searching, their approach effectively addresses both shortcut and short branches combination problems, particularly in complex tubular tree structures or challenging background environments.

Despite their notable improvements in tracking accuracy, nonlocal segment proposals-based grouping models~\cite{liu2021trajectory,wang2013interactive} still exhibit inherent limitations: 1) Existing approaches employing Dijkstra shortest path algorithm require complete graph information, necessitating weight computation for all edges. Given the non-negligible computational cost of calculating distances between segments—particularly when using advanced curvature-penalized geodesic distances—the process becomes computationally intensive and potentially redundant for large-scale graphs. 2) Since pre-extracted segment proposals serve as candidate nodes during graph construction, comprehensive prior knowledge is essential to establish connections that include valid paths from source to destination. Without such knowledge, the path searching algorithm may fail to incorporate the true minimal path in its solution set, resulting in shortcuts or complete path discovery failure. In Fig.~\ref{fig:example_minimal_path}, we illustrate the proposed method on a retinal image shown in~Fig.~\ref{fig_example_rawimage} and provides a comparison with other approaches, where the segment proposals are exhibited in Fig.~\ref{fig_example_proposals}. Fig.~\ref{fig_example_completegraph} shows a large graph used in~\cite{liu2021trajectory,wang2013interactive}, while Fig.~\ref{fig_example_dynamicgraph} the graph used in our model exhibiting significant reduction on the size of the graph. Figs.~\ref{fig_example_pointwise1} to~\ref{fig_example_proposed} illustrate the tubular centerline tracking results, proving the advantages of the proposed model. 

 In this paper, we propose a reinforcement learning-based tubular centerline  tracking model, named Dynamic Segment-proposal Grouping via Reinforcement Learning model (DSG-RL),  that employs Q-Learning and  segment graphs to seek shortest paths. The advantages of the proposed model are threefold. Firstly, our model enables localized graph exploration, eliminating the need to pre-compute all edge weights; only the weights of explored nodes and their selected neighbors require computation, substantially reducing computational complexity. Secondly, through an action-space incremental strategy, our method adaptively adjusts the exploration space for each segment, thereby obviating the reliance on comprehensive prior knowledge to the completeness of the solution set. Finally, comprehensive evaluations across three distinct tubular structures (vessels, roads, and rivers) demonstrate that our model indeed outperforms state-of-the-art point-wise and segment-wise methods in both accuracy and computational efficiency. 

%The main contributions of the paper are summarized as follows:
%\begin{itemize}
%	\item We establish a new framework for tracking tubular  centerlines via reinforcement learning and perceptual grouping. 
%	by formulating this problem tracking problem on tubular structures within the Markov decision process framework, offering a novel reinforcement learning-based solution paradigm.
%	\item We propose a dynamic segment-proposal grouping method leveraging Q-Learning that incorporates adaptive graph construction and segment-based structure tracking for optimal path identification.
%	\item Comprehensive evaluations across three distinct tubular structures—retinal vessels, roads, and rivers—demonstrate that our approach significantly outperforms state-of-the-art point-wise and segment-wise methods in both accuracy and computational efficiency.
%\end{itemize}

The paper is organized as follows. Section \ref{sec:background} revisits the foundation of curvature-penalized geodesic models and Q-Learning. Section \ref{sec:method} introduces our reinforcement learning-based segment grouping framework, including dynamic graph construction and adaptive exploration mechanisms. Section \ref{sec:experiment} presents comprehensive experiments across three diverse tubular structure datasets. Finally, Section \ref{sec:conclusion} concludes the paper with a summary of contributions and potential future directions.

\section{Background}
\label{sec:background}
This section presents the foundational components of our proposed methodology: the curvature-penalized geodesic model for distance evaluation between image points, and the Q-learning algorithm for optimal decision-making.

\subsection{Euler-Mumford Elastica Minimal Path Model}
\label{sec:curvature-penalized-geodesics}
The geodesic minimal path model computes the globally minimal geodesic distance by solving an Eikonal PDE~\cite{cohen1997global}.
The curvature-penalized geodesic models, such as the sub-Riemannian model~\cite{duits2018optimal} and the Euler-Mumford elastica model~\cite{chen2017global}, represent state-of-the-art approaches in geodesic modeling, yielding paths characterized by high degrees of smoothness and structural rigidity.

\textbf{Orientation lifting}.
The curvature-penalized geodesic models are defined on an orientation-lifted space $\bM=\Omega \times \bS^{1}$, where $\Omega \subset \bR^{2}$ is the open bounded image domain and $\bS^{1}=[0, 2\pi[$ is the angle space with a periodic boundary condition.
The problem is to determine the minimal geodesic in the lifted space $\bM$. Then a planar point $x$ in the image domain $\Omega$ can be regarded as a projection of the oriented-lifted point $\fx=(x, \theta)$ in the lifted space $\bM$. One can define $\bE=\bR^2\times\bR$ as the tangent space of any base point $\fx\in\bM$, where an element of $\bE$ is denoted as $\dfx=(\dx,\dtheta)$, with $\dx\in\bR^2$ and $\dtheta\in\bR$.
Furthermore, let $\gamma: [0,1] \to \Omega$ be a regular curve with non-vanishing velocity and $\tgamma =(\gamma,\eta):[0,1] \to \bM$ be its canonical orientation lifting, where $\gamma$ is referred to as the physical projection of the orientation-lifted path $\tgamma$ whose turning angles are represented by $\eta:[0,1]\to\bS^1$.

 \textbf{HJB PDE associated to the Elastica Model}.
The bending energy defined in the Euler-Mumford elastica model is defined as:
\begin{equation}
\label{eq_BendingEnergy}
\mathcal{L}(\tgamma)=\int_{0}^{1}\left (1+(\xi\kappa(t))^{2} \right )\|\gamma^\prime(t)\|\;dt,
\end{equation}
where $\kappa(t)=\eta'(t)/\| \gamma'(t) \|$ is the curvature of the planar curve $\gamma$ and $\xi$ is a positive constant that controls the related importance of the curvature term. The bending energy~\eqref{eq_BendingEnergy} corresponds to a unique Finsler geodesic metric $\cF:(\fx,\dfx)\in\bM\times\bE\mapsto[0,\infty]$, reading for any point $\fx=(x,\theta)\in\bM$ and any vector $ \dfx = (\dx, \dtheta) \in \bE$ as~\cite{chen2017anisotropic}
\begin{equation}
\label{eq_ElasticaMetric}
\cF(\fx,\dfx) = \left(1 + \frac{(\xi\dtheta)^2}{\|\dx\|^2}\right) \|\dx\|,
\end{equation}
if $\dx=\|\dx\|(\cos\theta,\sin\theta)$, i.e., the physical speed $\dx$ is positively aligned with the direction $(\cos\theta,\sin\theta)$. By convention $\cF(\fx,\mathbf{0})=0$ and in any other case $\cF(\fx,\dfx)=\infty$.

Given $ \fs\in \bM$ as a fixed source point, the global minimization of the weighted curve length~\eqref{eq_BendingEnergy} relies on the tool of geodesic distance map $\cU_{\fs}: \bM \to \bR$, which is defined by
\begin{align}
\cU_{\fs}(\fx)= \inf \{\mathcal{L}(\tgamma)~|~&\tgamma \in \mathrm{Lip}([0,1], \bM),\nonumber\\
&\tgamma(0) = \fs,\,\tgamma(1) = \fx\},
\end{align}
where $\mathrm{Lip}([0,1], \bM)$ denotes the space of Lipschitz-continuous curves $\gamma:[0,1]\to\bM$. It is known that the geodesic distance map $\cU_{\fs}$ satisfies the HJB equation with a boundary condition $\cU_{\fs}(\fs) = 0$ and  
\begin{equation}
\label{eq:eikonal}
\max_{\dfx \neq 0} 
\frac{\langle \nabla \cU_{\fs}(\fx),\dfx \rangle}
     {\cF(\fx,\dfx)} 
= 1,
\end{equation}
where $\nabla\cU_{\fs}$ is the standard Euclidean gradient of $\cU_{\fs}$. The HJB equation~\eqref{eq:eikonal} must hold at each point $\fx\in\bM\backslash\{\fs\}$ with  outflow boundary conditions on $\partial\bM$. The geodesic distance map $\cU_{\fs}$ is a viscosity solution to this HJB equation, accommodating potential singularities at the source point $\fs$.

The fast marching method~\cite{mirebeau2018fast} can be employed to solve a numerical approximation of the HJB equation~\eqref{eq:eikonal}. Subsequently, a geodesic path $\cC_{\fs, \fx}$ traveling from the source point $\fs$ to a target point $\fx$ can be obtained by re-parameterizing  the solution to a gradient descent ordinary differential equation (ODE) on $\cU_{\fs}$:
\begin{equation*}
\cC^\prime(u)
= -\arg\max\{ \langle\dfx, \nabla \cU_{\fs}(\cC(u))\rangle|\dfx\in\bE,\cF(\cC(u),\dfx)=1\}
\end{equation*}
till the source point $\fs$ is reached. 
Note that the solution $\cC$ to the gradient descent ODE is a geodesic path that links from $\fx$ to the source point $\fs$.

\subsection{Q-Learning}
Q-Learning is a foundational model-free reinforcement learning algorithm~\cite{shakya2023reinforcement} that enables an agent to learn an optimal policy for sequential decision-making within a Markov Decision Process (MDP) framework~\cite{puterman1990markov}. An MDP is formally defined by the tuple $(S, A, P, R, \beta)$, where:
\begin{itemize}
    \item $S$ is a finite set of states.
    \item $A$ is a finite set of actions.
    \item $P(s' \mid s, a)$ is the probability of transitioning from state $s \in S$ to state $s' \in S$ after taking action $a \in A$.
    \item $R(s, a, s')$ is the immediate reward received after transitioning from state $s$ to state $s'$ as a result of action $a$. Often, this is simplified to $R(s,a)$ if the reward only depends on the state and action, or $r_{t+1}$ denotes the specific reward sample received at time $t+1$.
    \item $\beta \in [0, 1[$ is the discount factor, which balances the importance of immediate versus future rewards.
\end{itemize}

The primary goal of Q-Learning is to find the optimal state-action value function, $Q^*(s, a)$. This function represents the expected cumulative discounted reward an agent can obtain by taking action $a$ in state $s$ and subsequently following the optimal policy $\pi^*$ thereafter:
\begin{equation}
    Q^*(s, a) = \mathbb{E}_{\pi^*} \left[ \sum_{k=0}^\infty \beta^k r_{t+k+1} \;\middle|\; s_t = s, a_t = a \right],
\end{equation}
where $r_{t+k+1}$ is the reward received at step $k+1$ in the future, given $s_t=s$ and $a_t=a$ at the current time $t$, and following the optimal policy. Q-learning iteratively estimates $Q^*(s,a)$ using a value $Q(s,a)$.

The learning process involves updating the current estimate $Q(s_t, a_t)$ for a given state-action pair $(s_t, a_t)$ based on the experienced transition $(s_t, a_t, r_{t+1}, s_{t+1})$, where $s_{t+1}$ is the next state and $r_{t+1}$ is the reward received. This update utilizes the Bellman optimality equation through the temporal difference (TD) error, $\delta_t$:
\begin{equation}
    \delta_t = r_{t+1} + \beta \max_{a'} Q(s_{t+1}, a') - Q(s_t, a_t).
\end{equation}
The Q-value for the state-action pair $(s_t, a_t)$ is then updated as:
\begin{equation}
    Q(s_t, a_t) \leftarrow Q(s_t, a_t) + \alpha \delta_t,
\end{equation}
where $\alpha \in ]0, 1]$ is the learning rate. This update rule incrementally adjusts $Q(s_t, a_t)$ towards the optimal value by minimizing the TD error, incorporating the observed reward $r_{t+1}$ and the estimated maximum Q-value for the next state $s_{t+1}$.

To manage the exploration-exploitation trade-off, Q-Learning commonly employs an $\epsilon$-greedy strategy~\cite{wiering2012reinforcement}. With probability $\epsilon$, the agent chooses a random action to explore the environment, and with probability $1-\epsilon$, it selects the action that maximizes the current Q-value (exploitation). Typically, $\epsilon$ is annealed (gradually decreased) over time to favor exploitation as learning progresses.

Under standard conditions, including sufficient exploration of all state-action pairs and an appropriate learning rate schedule, Q-Learning is proven to converge to the optimal Q-function, $Q^*(s, a)$~\cite{sutton1998reinforcement}. This guarantee of optimality makes Q-Learning a powerful tool for problems with discrete state and action spaces, such as the tubular structure tracking problem formulated in this work, where finding a globally optimal path is the primary objective.

\section{The DSG-RL Model}
\label{sec:method}
This section details the proposed DSG-RL for detecting and establishing connectivity within tubular structures. We begin with an overview of the entire framework before elaborating on its core components.

\subsection{Overview of DSG-RL}

\begin{figure}[t]
\centering
\includegraphics[width=\columnwidth]{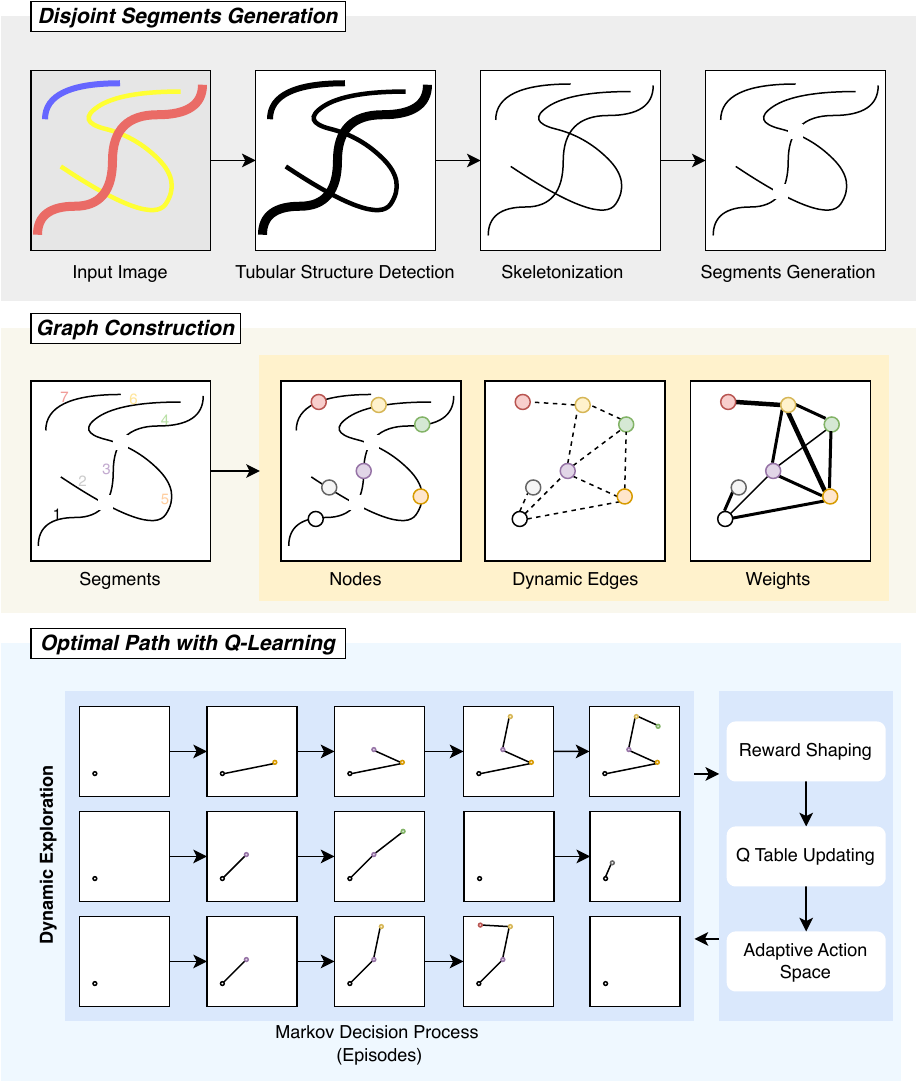}
\caption{The framework of the proposed DSG-RL model. }
\label{fig:framework}	
\end{figure}
 
Fig.~\ref{fig:framework} illustrates the architecture of the proposed DSG-RL model. The method operates on an abstract graph where nodes represent centerline segments of the tubular structure. Instead of pre-computing a complete, static graph, our approach discovers edges dynamically, weighting them with the geometric properties of curvature-penalized minimal paths. A Q-Learning agent is then employed to learn a policy for navigating this partially-observed graph, identifying an optimal path that ensures both structural integrity and global coherence of the final tracked tubular structure.

\begin{figure*}[htb]
\centering
\includegraphics[height=0.25\textwidth]{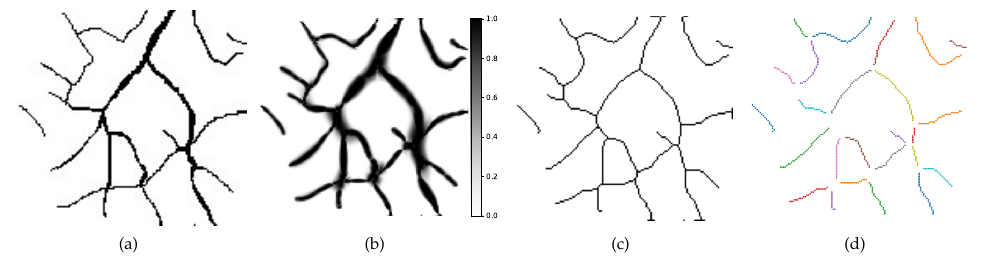}
\caption{Visualization of the generation of a set of disjoint segments. (\textbf{a})~An image that contains complex tubular structures. (\textbf{b})~Visualization of the vesselness map that indicates the probability that a point belongs to a tubular structure. (\textbf{c})~The skeleton structure of the tubular structures. (\textbf{d})~Visualization of the generated disjoint segments where all the junction points in the skeleton map are removed.}
\label{fig:DisjointSegmentsGeneration}             
\end{figure*}

\begin{figure*}[t]
 \includegraphics[width=\textwidth]{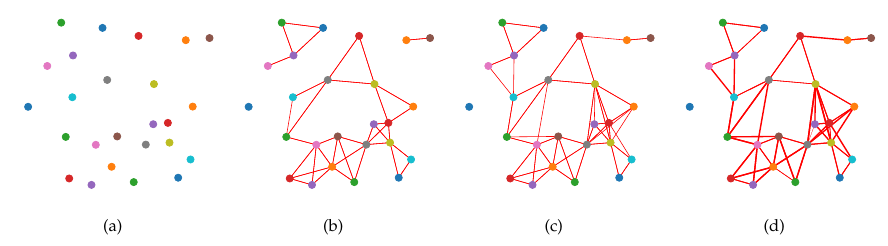}
\caption{Illustration of the graph representation derived from disjoint segments. (\textbf{a})~Nodes corresponding to disjoint segments, colored for identification. (\textbf{b})~Associated graph structure with a short extension length ($\ell=1$). (\textbf{c})~A more densely connected graph resulting from a larger extension length ($\ell=10$). (\textbf{d})~The weighted graph for the $\ell=10$ case, where edge widths are proportional to their weights.}
\label{fig:GraphConstruction}             
\end{figure*}

\subsection{Generation of Tubular Structure Centerlines}

In this work, a centerline segment of a tubular structure is defined as a regular curve and can be derived by computing  the medial axis or skeleton of these elongated shapes. In practice, an image usually contains complex tubular structures of multiple branches and topologies, and as a consequence the connection between two centerline segments is sometimes unreliable. Therefore,  we consider to extract a set of disjoint centerline segments, denoted by $T_i\subset\Omega$ and indexed by $1\leq i \leq M$ with $M$ a positive integer. A target tubular structure is therefore composed of an ordered sequence of these disjoint centerline segments.

The numerical implementation for constructing these disjoint centerline segments can be carried out via a segmentation-then-skeletonization scheme, as considered in~\cite{liu2021trajectory,xu2011vessel}. This scheme is efficient since it can take advantages of existing  approaches to generate the tubular structure segmentation results. For the sake of simplicity, we apply the oriented flux filter to enhance the image data, yielding a vesselness map. Then a thresholding value is utilized to generate the binary segmentation. Following a skeletonization procedure and removing all the branch points, we obtain the target disjoint centerline segments $T_i\subset\Omega$.
 In Fig.~\ref{fig:DisjointSegmentsGeneration}, we illustrate the procedure of the generation of $T_i\subset\Omega$.

\subsection{Graph Representation and Edge Weighting}\label{sec:graph_construction}

Each generated segment $T_i$ is mapped to a node $v_i$ in a graph $G = (V, E)$, where $V = \{v_1, \ldots, v_M\}$ is the set of all segment nodes and $E$ is the set of edges discovered by the learning agent. Edges in $G$ are not pre-computed but are established on-demand during the exploration phase.

The discovery of potential edges from a node $v_i$ is initiated by extending its corresponding segment, $T_i$, from its endpoints ($p_{i,0}, p_{i,1}$) along their tangent directions. The length of this extension, $\ell$, is dynamically sampled to probe for connections at various scales, as detailed in Section~\ref{sec:method_exploration}. A tubular patch, $P_i$, is then defined around the extended segment $\tilde{T}_i$. An edge $(v_i, v_j)$ is considered a candidate for exploration if the segment $T_j$ intersects with the patch $P_i$. This dynamic process allows the graph structure to evolve as the agent explores, as illustrated conceptually for different extension lengths in Fig.~\ref{fig:GraphConstruction}.

The weight $\omega_{i,j}$ assigned to a candidate edge $(v_i, v_j)$ is defined as the geodesic distance between the corresponding segments $T_i$ and $T_j$. For each pair of segments, we first identify the optimal connection points by computing the Euclidean distances between all pairs of endpoints and selecting the pair $(p_i, p_j)$ with minimal distance, where $p_i \in \{p_{i,0}, p_{i,1}\}$ and $p_j \in \{p_{j,0}, p_{j,1}\}$.

To incorporate curvature-aware path metrics, we employ the Finsler metric $\cF$, as formulated in Eq.~\eqref{eq_ElasticaMetric}, to compute the geodesic distance. The process begins by embedding the endpoints $p_i$ and $p_j$ into the orientation-lifted space $\bM = \Omega \times \mathbb{S}^1$, where each point is augmented with an orientation $\theta \in \mathbb{S}^1$. The resulting lifted points are denoted as $\fp_i = (p_i, \theta_i)$ and $\fp_j = (p_j, \theta_j)$, where $\theta_i$ and $\theta_j$ represent the tangent orientations of the segments $T_i$ and $T_j$ at the respective endpoints.
The geodesic distance is computed by solving the HJB PDE~\eqref{eq:eikonal} with the Finsler metric $\cF$ using an efficient Hamiltonian fast marching method~\cite{mirebeau2018fast}. The resultant distance value $\cU_{\fp_i}(\fp_j)$ quantifies the minimal path energy between the segments, incorporating both spatial proximity and curvature penalties. An example of the resulting weighted graph structure is illustrated in Fig.~\ref{fig_GraphConstruction_wg}.

\subsection{Optimal Path with Q-Learning}
\begin{figure}[htbp]
\centering
\includegraphics[width=\columnwidth]{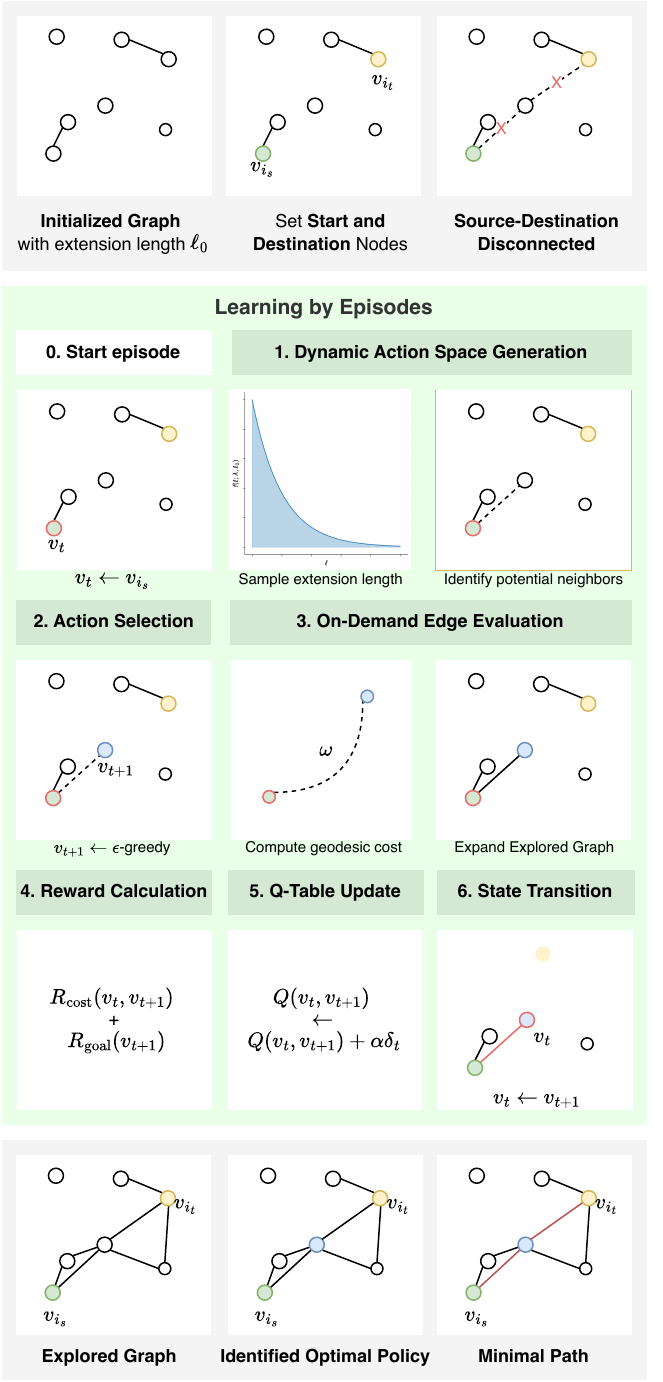}
\caption{The minimal path tracking process via Q-Learning. The process starts with an initial graph that may not connect the source ($v_{is}$) and destination ($v_{it}$) nodes (top). The agent then learns through episodes (middle), where it iteratively: (1) generates a dynamic action space, (2) selects an action, (3) evaluates the new edge on-demand, (4) calculates a reward, (5) updates the Q-table, and (6) transitions to a new state. This allows the agent to expand its knowledge of the graph and ultimately extract the optimal path (bottom).}
\label{fig:minimal_path_q_learning}	
\end{figure}

\subsubsection{Tubular Structure Tracking by Segments Grouping}
We formulate the tubular structure tracking problem within a weighted graph $G = (V, E)$, constructed based on segments as detailed in Section~\ref{sec:graph_construction}. Each segment corresponds to a node in $V$, while the edges $E$ represent the spatial relationships between segments.

Given a source point $p_s$ and a destination point $p_t$, we identify their corresponding segments, $T_{i_s}$ and $T_{i_t}$, and the associated graph nodes, $v_{i_s}$ and $v_{i_t}$. 
The objective is to find an optimal path---a sequence of connected nodes---from $v_{i_s}$ to $v_{i_t}$. This optimal path, $\gamma^* = (v^*_{(1)}, v^*_{(2)}, \ldots, v^*_{(L^*)})$, is the one that minimizes the cumulative sum of edge weights:
\begin{equation}
	\gamma^* = \underset{\gamma \in \Pi(v_{i_s}, v_{i_t})}{\arg\min}\;\sum_{l=1}^{L-1} \omega_{(l),(l+1)},
\end{equation}
where $\Pi(v_{i_s}, v_{i_t})$ is the set of all feasible paths from $v_{i_s}$ to $v_{i_t}$. A path $\gamma = (v_{(1)}, \ldots, v_{(L)})$ is a sequence where $v_{(1)} = v_{i_s}$, $v_{(L)} = v_{i_t}$, and each pair $(v_{(l)}, v_{(l+1)})$ constitutes a valid edge in $E$. The weight $\omega_{(l),(l+1)}$ quantifies the geodesic distance between the corresponding segments $T_{(l)}$ and $T_{(l+1)}$.

\subsubsection{Mapping to Markov Decision Process}
We formulate the minimal path problem as an MDP by augmenting the state to encapsulate the agent's dynamically acquired knowledge of the graph structure. This ensures the MDP remains stationary, with fixed state and action spaces, accommodating our adaptive exploration strategy (Section~\ref{sec:method_exploration}). The MDP is defined by the tuple $(S, A, P, R, \beta)$, where the components are specified as follows:

\begin{itemize}
    \item \textbf{State Space ($S$):} A state $s \in S$ is a tuple $s = (v, \mathcal{E})$, where $v \in V$ is the agent's current node (segment) and $\mathcal{E} \subseteq V \times V$ is the set of all graph edges discovered by the agent up to the current time. In practice, this is implemented not by indexing a Q-table by $\mathcal{E}$, but by dynamically expanding the table as new edges are found (see Section~\ref{sec:update_with_dynamic_actions}).

    \item \textbf{Action Space ($A$):} The action space $A$ is the universal set of all nodes in the graph, i.e., $A=V$. For any state $s=(v, \mathcal{E})$, the set of permissible actions is $A(s) = \{v' \in V \mid (v, v') \in \mathcal{E}\}$.

    \item \textbf{Transition Function ($P$):} The state transition captures both the agent's movement and the discovery of new graph connections. Upon taking an action $a_t = v' \in A(s_t)$ from state $s_t = (v_t, \mathcal{E}_t)$, the system transitions to a new state $s_{t+1} = (v_{t+1}, \mathcal{E}_{t+1})$. Here, $v_{t+1} = v'$, and the updated edge set $\mathcal{E}_{t+1}$ is formed by the union of $\mathcal{E}_t$ and any new edges discovered from node $v'$ via the stochastic exploration mechanism detailed in Section~\ref{sec:method_exploration}. This formulation preserves the Markov property, as $s_{t+1}$ depends only on $s_t$ and $a_t$.

    \item \textbf{Reward Function ($R$):} The reward $R(s_t, a_t)$ is determined by the transition from node $v_t$ to $v_{t+1}$ and is independent of the knowledge component $\mathcal{E}_t$. It comprises a cost proportional to the geodesic distance and a terminal bonus for reaching the target, as defined in Section~\ref{sec:reward_shaping}.

    \item \textbf{Discount Factor:} $\beta \in [0, 1[$ is the standard discount factor.
\end{itemize}

This formulation grounds our method in standard reinforcement learning framework, enabling the application of Q-Learning to find an optimal policy $\pi^*: S \to A$. The objective is to learn optimal state-action value function, $Q^*(s, a)$, which represents the maximum expected cumulative reward for taking action $a$ in state $s$ and following $\pi^*$ thereafter. The value function $Q^*(s, a)$ satisfies the Bellman's optimality equation:
\begin{equation}
	Q^*(s_t, a_t) = R(s_t, a_t) + \beta \, \mathbb{E}\left[\max_{a' \in A(s_{t+1})} Q^*(s_{t+1}, a')\right]
\end{equation}
where the expectation accounts for the stochasticity in the state transition, including the potential discovery of new edges. The learned policy $\pi^*$ directs the agent from the source node $v_{i_s}$ to the target $v_{i_t}$ along the path of minimum cumulative cost.

\subsubsection{Dynamic Exploration via Adaptive Action Space}\label{sec:method_exploration}
A critical challenge in segment-wise graph search arises when the initial graph construction fails to include essential connections between the source point $p_s$ and destination point $p_t$, potentially preventing the discovery of a valid path. This typically occurs if the segment extension length $\ell$, used to define segment neighborhoods (Section~\ref{sec:graph_construction}), is insufficient. To address this, we introduce a dynamic exploration strategy that expands the set of known edges on-the-fly as the agent traverses the graph.

Our method discovers connections in an on-demand fashion. At each state $s_t=(v_t, \mathcal{E}_t)$, the agent generates a dynamic action space $A(s_t)$ by probing the neighborhood of its current segment $T_t$. This process begins by sampling an extension length $\ell$ from a shifted exponential probability density function (PDF):
\begin{equation}\label{eq:length_pdf}
  f(\ell; \lambda, \ell_0) = \lambda e^{-\lambda(\ell - \ell_0)} \mathbf{1}_{[\ell_0, \infty)}(\ell),
\end{equation}
where $\ell_0 > 0$ is a minimum extension length and $\lambda > 0$ is the rate parameter. This stochastic sampling allows the agent to probe for connections at varying distances. Using the sampled $\ell$, a set of potential new neighboring segments, $N(v_t)$, is identified as described in Section~\ref{sec:graph_construction}. The action space for the current step, $A(s_t)$, is then the union of already-known neighbors from $\mathcal{E}_t$ and these newly identified potential neighbors in $N(v_t)$.

From this dynamically generated action space, an action $a_t=v_{t+1}$ is selected using an $\epsilon$-greedy policy:
\begin{equation}\label{eq:node_selection}
    a_t = \begin{cases} 
    \text{random action from } A(s_t) & \text{w.p. } \epsilon \\ 
    \arg\max_{a' \in A(s_t)} Q(s_t, a') & \text{w.p. } 1-\epsilon 
    \end{cases}
\end{equation}
where the Q-value $Q(s_t, a')$ for a new, potential action is initialized to an optimistic value to encourage exploration. Crucially, if the selected action corresponds to traversing a new, previously unknown edge (i.e., $(v_t, v_{t+1}) \notin \mathcal{E}_t$), only then is the geodesic cost for this specific edge computed. This "lazy evaluation" is the key to our method's computational efficiency. The new edge and its weight are then added to the knowledge base, forming the updated set $\mathcal{E}_{t+1} = \mathcal{E}_t \cup \{(v_t, v_{t+1})\}$. This on-demand computation focuses computational effort only on paths the agent actively explores. After executing $a_t$, the agent transitions to the new state $s_{t+1}=(v_{t+1}, \mathcal{E}_{t+1})$, and this stochastic discovery process defines the transition probabilities $P(s_{t+1} | s_t, a_t)$ of our MDP.

\subsubsection{Reward Shaping for Optimal Path Guidance}\label{sec:reward_shaping}
To guide the agent's learning process within the MDP framework, we define a reward function $R(s_t, a_t)$ that incentivizes both path efficiency and successful goal attainment. As established in our MDP formulation, the reward for taking action $a_t$ in state $s_t=(v_t, \mathcal{E}_t)$ depends only on the transition from the current node $v_t$ to the next node $v_{t+1}$, which is determined by the action $a_t$. The immediate reward, $r_{t+1} = R(s_t, a_t)$, is composed of a transition cost and a terminal bonus.

\textbf{Transition Cost.} Traversing from node $v_t$ to $v_{t+1}$ incurs a cost proportional to the geodesic distance between their corresponding segments. This cost is represented by the negative of the edge weight $\omega(v_t, v_{t+1})$:
\begin{equation}
	R_{\text{cost}}(v_t, v_{t+1}) = -\omega(v_t, v_{t+1}).
\end{equation}
This penalty encourages the agent to find paths with minimal cumulative geodesic distance.

\textbf{Goal Attainment Bonus.} A substantial positive reward, $\mathcal{G} > 0$, is awarded if the selected action $a_t$ leads the agent to the target node $v_{i_t}$. This terminal bonus is defined as:
\begin{equation}
	R_{\text{goal}}(v_{t+1}) =
\begin{cases}
\mathcal{G}, & \text{if } v_{t+1} = v_{i_t} \\
0, & \text{otherwise}.
\end{cases}
\end{equation}
This component provides a strong incentive for the agent to complete the tracking task.

The total immediate reward for the transition is the sum of these components:
\begin{equation}
	R(s_t, a_t) = R_{\text{cost}}(v_t, v_{t+1}) + R_{\text{goal}}(v_{t+1}).
\end{equation}
The Q-Learning algorithm seeks to find a policy that maximizes the expected cumulative discounted reward, $\mathbb{E}\left[\sum_{k=0}^\infty \beta^k r_{t+k+1}\right]$. This reward structure effectively guides the agent toward paths that are both geometrically efficient and successfully connect the source to the target.

\subsection{Implementation Details}\label{sec:implementation_details}

\begin{algorithm}[t]
\caption{Main Algorithm of DSG-RL}
\label{alg:tubular_tracking}
\begin{algorithmic}[1]
\REQUIRE Image $\mathbf{I}$, source point $p_s$, destination point $p_t$.

\STATE \textbf{Step 1: Disjoint Segments Generation}
\STATE Generate a vesselness score map from $\mathbf{I}$ and apply skeletonization to derive a set of disjoint centerline segments $T = \{T_1, \ldots, T_M\}$.

\STATE \textbf{Step 2: Graph Initialization}
\STATE Create a node $v_i$ for each segment $T_i$ to form the set of vertices $V$.
\STATE Construct an initial graph $G = (V, E_0)$ using a minimum extension length $\ell_0$ to define the initial edge set $E_0$.
\STATE Identify the source node $v_{i_s}$ and destination node $v_{i_t}$ corresponding to $p_s$ and $p_t$.

\STATE \textbf{Step 3: Dynamic Path Discovery via Q-Learning}
\STATE Employ Q-Learning to find an optimal policy from $v_{i_s}$ to $v_{i_t}$.
\STATE \quad Explore the graph dynamically, computing geodesic edge weights only for connections visited by the agent.
\STATE From the learned policy, extract the optimal sequence of segments $\{v^*_{(l)}\}_{l=1}^{L^*}$.

\STATE \textbf{Step 4: Path Reconstruction}
\STATE Sequentially connect the segments in the optimal path $\{v^*_{(l)}\}$ by computing the geodesic minimal path between their consecutive endpoints.
\STATE Concatenate the segments and connecting paths to form the final path $\gamma$.

\RETURN The complete minimal path $\gamma$ from $p_s$ to $p_t$.
\end{algorithmic}
\end{algorithm}

\subsubsection{Q-Table Update with a Dynamic Action Space}\label{sec:update_with_dynamic_actions}
Unlike conventional Q-learning, which assumes a stationary action space, our DSG-RL framework operates in a non-stationary environment where the set of available actions evolves as new connections between segments are discovered. This problem is related to the broader challenge of lifelong or continual learning, where an agent must adapt to changing dynamics~\cite{chandak2020lifelong}. Our approach addresses this by employing an adaptive Q-table update mechanism.

Our method adopts a hybrid strategy for graph construction to balance efficiency and robustness. Initially, a minimal graph is constructed using a base extension length $\ell_0$, providing a foundational set of actions and preventing an inefficient cold start. During training, the agent discovers new actions by sampling larger extension lengths ($\ell > \ell_0$) that may reveal new segment connections. When such a connection is found, the Q-table is dynamically expanded to include the new state-action pair. Subsequent policy updates operate on this evolving Q-table, allowing the agent to integrate new topological information into its decision-making.

\subsubsection{Tubular Structure Reconstruction}
Upon convergence of the Q-Learning algorithm, the optimal policy $\pi^*$ delineates an optimal sequence of nodes, denoted as $\{v^*_{(l)}\}_{l=1}^{L^*}$, which connects the source node $v_{i_s}$ to the target node $v_{i_t}$. The reconstruction of the final continuous tubular path involves bridging the discontinuities between consecutive segments within this optimal sequence. For each adjacent pair of segments $(T_{(l)}, T_{(l+1)})$, corresponding to the nodes $(v^*_{(l)}, v^*_{(l+1)})$ in the optimal sequence, their respective nearest endpoints are identified. Subsequently, a geodesic minimal path is computed to connect these endpoints within the continuous image domain using the Finsler metric $\cF$. The concatenation of the original constituent segments with these computed inter-segment geodesic paths yields the final, globally coherent tubular structure track. This reconstructed path inherently adheres to the underlying geometry of the structure and preserves smoothness characteristics.

\subsubsection{Computational Complexity}
The primary computational advantage of DSG-RL stems from its on-demand approach to graph construction, which circumvents the high cost of pre-computing a complete graph required by traditional segment-wise methods. Conventional approaches, which typically employ algorithms like Dijkstra's, must first build a graph with $M$ nodes (segments) and a set of edges $E$. This requires computing the geodesic distance for all $|E|$ edges—the main computational bottleneck. Subsequently, running Dijkstra's algorithm adds a complexity of $O(|E| + M \log M)$.

In stark contrast, our DSG-RL method does not require a pre-built, fully-weighted graph. Instead, it explores the graph dynamically, calculating an edge's geodesic weight only at the moment it is first traversed by the learning agent. Because the goal-directed policy guides the agent to explore promising regions of the graph, only a small fraction of the total possible edges are ever evaluated. This on-demand evaluation strategy drastically reduces the number of expensive geodesic computations, leading to significant efficiency gains without sacrificing path quality, particularly in dense graphs where $|E|$ can be large.

\section{Experimental Results}
\label{sec:experiment}

\begin{figure*}[htbp] 
\centering
\includegraphics[width=0.90\textwidth]{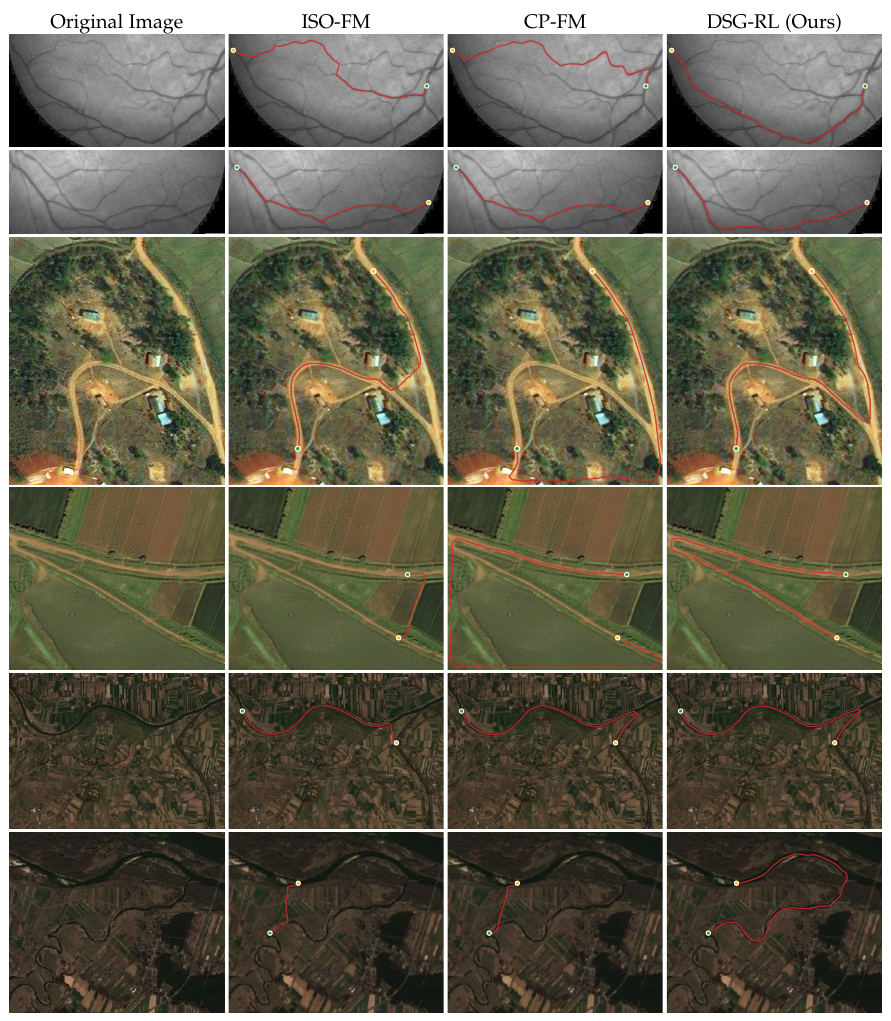}
\caption{Qualitative comparison of various tubular structure tracking methods. The columns, from left to right, depict results from: Isotropic Fast Marching (ISO-FM), Curvature Penalized Fast Marching (CP-FM), and the proposed Segment Grouping with Reinforcement Learning (DSG-RL) approach. The first two rows illustrate distinct test cases from the IOSTAR retinal vessel dataset; the subsequent two rows present cases from the DeepGlobe roads dataset; and the final two rows correspond to examples from the Sentinel River Dataset. In each subfigure, the green and yellow dots denote the designated start and end points, respectively, while the red line delineates the minimal path identified by the respective method.}
\label{fig:pixel_wise_comparison}
\end{figure*}

\begin{figure*}[htbp]
\centering
\begin{subfigure}[b]{0.3\textwidth}
    \includegraphics[width=\textwidth]{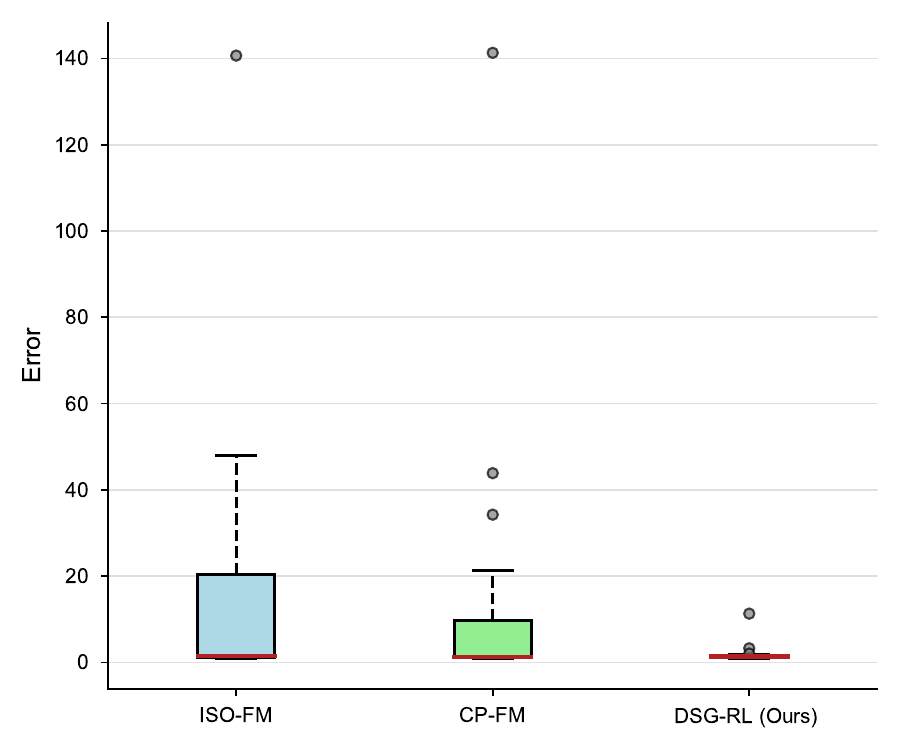}
    \caption{}
    \label{fig:acc_star}
\end{subfigure}
%\hfill % Distribute subfigures horizontally
\begin{subfigure}[b]{0.3\textwidth}
    \includegraphics[width=\textwidth]{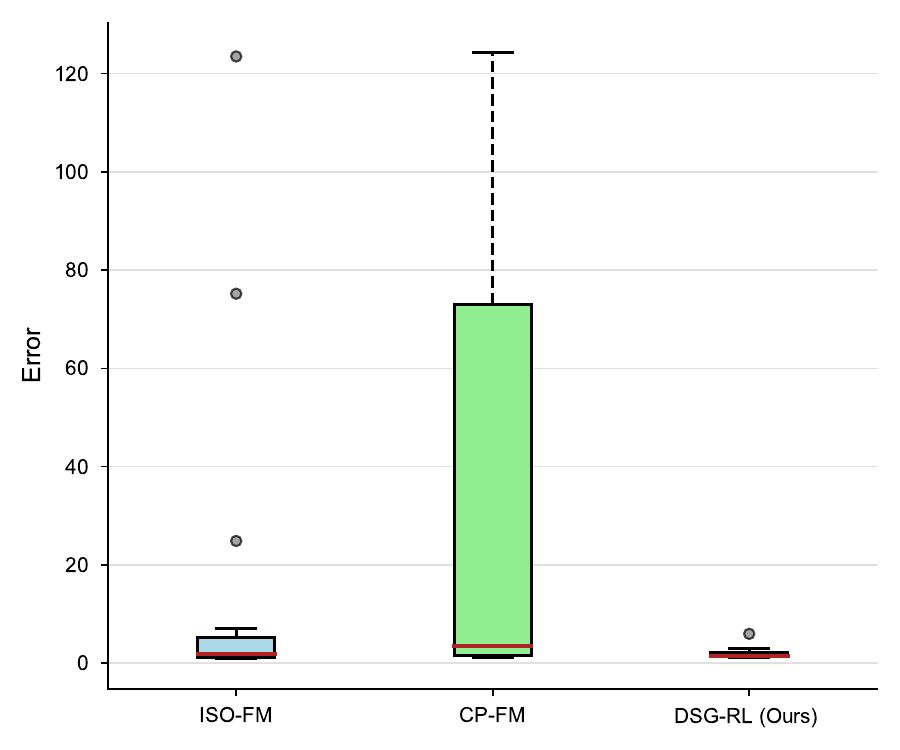}
    \caption{}
    \label{fig:acc_road}
\end{subfigure}
%\hfill % Distribute subfigures horizontally
\begin{subfigure}[b]{0.3\textwidth}
    \includegraphics[width=\textwidth]{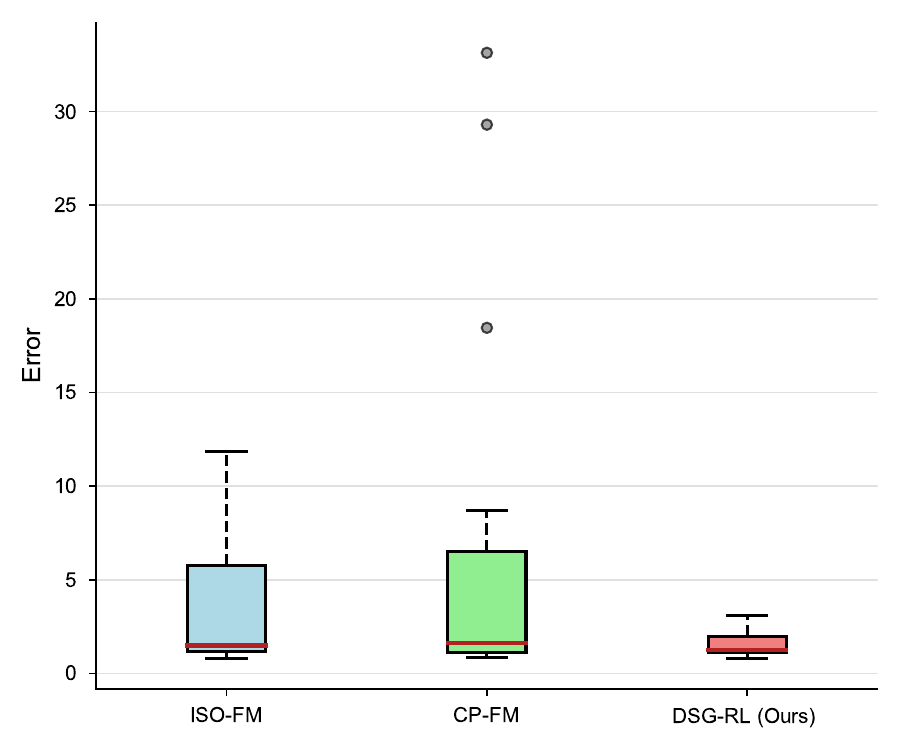}
    \caption{}
    \label{fig:acc_river}
\end{subfigure}
\caption{Quantitative evaluation of centerline error across different datasets. Boxplots illustrate the distribution of errors for each method. (\textbf{a})~IOSTAR Retinal Vessel dataset. (\textbf{b})~DeepGlobe Road dataset. (\textbf{c})~Sentinel River dataset.}
\label{fig:accuracy_pixel_wise}
\end{figure*}

\begin{figure*}[htbp] 
\centering
\includegraphics[width=0.8\textwidth]{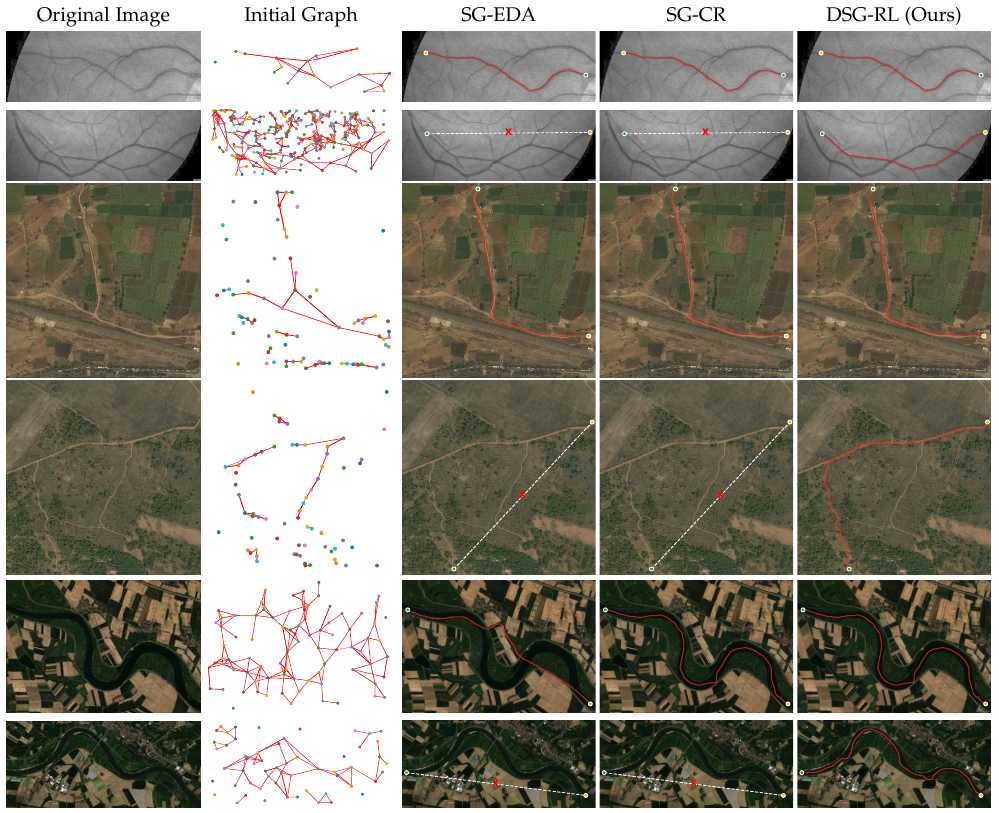} 
\caption{Qualitative comparison of segment-wise tubular structure tracking method. From left to right, columns display: the original image, the generated graph, results from Segment Grouping with Euclidean Distance and Angles (SG-EDA), Segment Grouping with Curvature Regularization (SG-CR), and our proposed Segment Grouping with Reinforcement Learning (DSG-RL) method. The first two rows showcase test cases from the IOSTAR retinal vessel dataset, the next two rows from the DeepGlobe roads dataset, and the final two rows from the Sentinel River Dataset. For each dataset, the first example demonstrates a dense graph configuration (large extension length $\ell$), while the second illustrates a sparse graph configuration (small extension length $\ell$). In each subfigure, green and yellow markers denote the start and end points, respectively; the red line indicates the identified minimal path. A dashed line connects the start and end points if no valid path is found.}
\label{fig:segment_wise_comparison}
\end{figure*}

This section presents a comprehensive evaluation of the proposed DSG-RL method through qualitative and quantitative comparisons with state-of-the-art approaches across three distinct tubular structure datasets.

\subsection{Experimental Setup}
\subsubsection{Datasets}
Our evaluation utilizes three distinct datasets to ensure a thorough assessment across different domains:
\begin{itemize}
	\item \textbf{IOSTAR Retinal Vessel Dataset}~\cite{7530915}: This dataset contains 1024$\times$1024 pixel retinal images with expert-annotated vessel structures, providing a benchmark for medical imaging applications.
	
	\item \textbf{DeepGlobe Road Dataset}~\cite{DeepGlobe18}: This dataset consists of 1024$\times$1024 pixel high-resolution (50 cm/pixel) satellite images from diverse geographic locations, designed for road extraction tasks.
	
	\item \textbf{Sentinel River Dataset}\footnote{https://github.com/radekszostak/sentinel-river-segmentation-dataset}: This dataset includes 400$\times$400 pixel Sentinel II satellite images of river systems, offering challenges related to natural water body delineation.
\end{itemize}

\subsubsection{Baseline Methods}

We compare our DSG-RL method against both point-wise and segment-wise state-of-the-art techniques in tubular structure tracking:

\begin{itemize}
	\item \textbf{Isotropic Fast Marching (ISO-FM)}~\cite{cohen1997global}: A classical point-wise method that computes minimal paths based on a Riemannian metric.
	
	\item \textbf{Curvature Penalized Fast Marching (CP-FM)}~\cite{mirebeau2018fast}: An advanced point-wise method that incorporates curvature penalization to produce smoother, more realistic paths.
	
	\item \textbf{Segment Grouping with Euclidean Distance and Angles (SG-EDA)}~\cite{wang2013interactive}: A segment-wise approach where connection costs are based on Euclidean distance and angular alignment between segments.
	
	\item \textbf{Segment Grouping with Curvature Regularization (SG-CR)}~\cite{liu2021trajectory}: A sophisticated segment-wise method that uses curvature-penalized geodesic paths to connect segments within a globally optimal graph search.
\end{itemize}

\subsection{Comparison with Point-Wise Methods}

We first benchmark DSG-RL against the point-wise ISO-FM and CP-FM methods.
The qualitative results in Fig.~\ref{fig:pixel_wise_comparison} highlight the limitations of point-wise methods and the superior performance of our DSG-RL approach. Point-wise methods, particularly ISO-FM, are prone to shortcutting. For example, in the first retinal image, ISO-FM incorrectly jumps between vessel branches. This issue is more pronounced in the road and river examples, where both ISO-FM and CP-FM fail to follow the curvilinear structures, instead cutting across fields and river meanders. While CP-FM shows some improvement over ISO-FM by penalizing curvature, it still often fails to adhere to the true path, as seen in the road and river datasets. In contrast, our DSG-RL method consistently and accurately tracks the centerlines across all datasets. It successfully navigates the tortuous vessel paths, follows the winding roads, and traces the meandering rivers, demonstrating its robustness and effectiveness in handling diverse and complex tubular structures.

For quantitative analysis, we use the mean centerline error ($\varepsilon$) as the primary metric, defined as:
\begin{equation}
\label{eq:mean_centerline_error}
	\varepsilon = \frac{1}{N}  \sum_{x_i \in \gamma} \min_{x^{\text{gt}} \in \gamma^{\text{gt}}} \|x_i - x^{\text{gt}}\|,
\end{equation}
where $\gamma$ is the predicted path with $N$ points, $\gamma^{\text{gt}}$ is the ground-truth path, and $\| \cdot \|$ is the Euclidean distance.

Fig.~\ref{fig:accuracy_pixel_wise} displays the quantitative results on a test set of 20 image patches per dataset. 
Our DSG-RL method consistently outperforms both point-wise baselines across all three datasets. The superior performance highlights the advantage of our segment-based approach, which leverages local structural coherence to avoid the errors common in point-wise tracking.

\subsection{Comparison with Segment-Wise Methods}
To demonstrate the advantages of our dynamic approach, we compare DSG-RL with the segment-wise SG-EDA and SG-CR methods, focusing on both accuracy and computational efficiency.

\subsubsection{Accuracy}
We evaluate the accuracy under two types of graph configurations: 1) a \textbf{dense graph}, constructed with a large value of  extension length $\ell$ to ensure all necessary connections are present, and 2) a \textbf{sparse graph}, using a low value of $\ell$ that may result in an incomplete set of connections.

Qualitative results are presented in Fig.~\ref{fig:segment_wise_comparison}. The findings reveal two key insights:
1) The choice of edge weight metric is crucial. As shown in the dense graph cases (e.g., first river example), the advanced geodesic metric used by SG-CR and our method yields more accurate paths than the simpler Euclidean and angular metrics of SG-EDA.
2) Static methods fail when the initial graph is incomplete. In sparse graph scenarios (second example for each dataset), both SG-EDA and SG-CR are unable to find a path because the necessary connections were not established during initial graph construction. Our DSG-RL method, with its dynamic exploration capability, successfully discovers the missing links and identifies the correct minimal path.

\subsubsection{Efficiency}
We assess computational efficiency by comparing the number of geodesic distance calculations—the most expensive step—required by our DSG-RL method and the state-of-the-art SG-CR.

\begin{table}[htbp]
	\centering
	\caption{Computational cost comparison between SG-CR and our proposed DSG-RL method. The values represent the average number of geodesic distance calculations required per pathfinding task. Cost Saved (\%) quantifies the percentage reduction in computations achieved by DSG-RL.}
	\label{tab:cost_comparison}
	\begin{tabular}{lccc}
		\toprule
		\textbf{Dataset} & \textbf{SG-CR} & \textbf{DSG-RL} & \textbf{Cost Saved (\%)} \\
		\midrule
		IOSTAR Retinal Vessel             & 319.92                     & 252.60                 & 21.04                    \\
		DeepGlobe Road             & 227.20                     & 145.50                 & 35.96                    \\
		Sentinel River            & 722.58                     & 567.53                 & 21.46                    \\
		\midrule
		Overall          & 410.48                     & 312.62                 & 23.84                    \\
		\bottomrule
	\end{tabular}
\end{table}

Table~\ref{tab:cost_comparison} summarizes the computational costs. DSG-RL achieves a significant reduction in geodesic computations across all datasets, with an overall cost saving of 23.84\%. This efficiency gain is a direct result of our "lazy evaluation" strategy. The reinforcement learning agent intelligently prioritizes promising connections, avoiding the exhaustive and often redundant computation of all possible edge weights required by SG-CR. This targeted exploration is particularly advantageous in complex networks with many segments. It is worth noting that these savings are demonstrated on small image patches; the computational advantage of DSG-RL is expected to be even more substantial on full-sized images containing a much larger number of segments, many of which are irrelevant to the desired path.

\section{Conclusion}
\label{sec:conclusion}
In this work we introduced a novel reinforcement learning-based dynamic segment-proposal grouping method, named the DSG-RL, for estimating accurate shortest paths to track the centerlines of tubular structures in complex scenarios. We addressed the inherent limitations of conventional segment-wise approaches, which are often hampered by high computational costs from exhaustive graph pre-computation and failures due to incomplete structural information. By formulating the tracking task as an MDP, our method employs Q-Learning to dynamically explore a graph of segments, computing connection costs on-demand and adaptively expanding the search space. This strategy not only circumvents the need for a fully pre-computed graph but also demonstrates remarkable resilience to sparse or incomplete initial data. Comprehensive evaluations on diverse datasets, including retinal vessels, road networks, and river systems, confirmed that DSG-RL significantly surpasses state-of-the-art point-wise and segment-wise methods in both tracking accuracy and computational efficiency. Our framework effectively preserves global path coherence across complex topologies without depending on extensive prior structural knowledge, establishing a new and efficient paradigm for tubular structure analysis.

%\section*{Acknowledgments}
%This research was supported in part by the National Natural Science Foundation of China (Grant No. 62206144), the Shandong Provincial Natural Science Foundation (Grant No. ZR2022QF042), the Taishan Scholar Program for Distinguished Experts (Grant No. tstp20250537), and the Key R\&D Program of Shandong Province (Major Scientific and Technological Innovation Project, Grant No. 2024CXGC010109).

%
%{\appendix[Proof of the Zonklar Equations]
%Use $\backslash${\tt{appendix}} if you have a single appendix:
%Do not use $\backslash${\tt{section}} anymore after $\backslash${\tt{appendix}}, only $\backslash${\tt{section*}}.
%If you have multiple appendixes use $\backslash${\tt{appendices}} then use $\backslash${\tt{section}} to start each appendix.
%You must declare a $\backslash${\tt{section}} before using any $\backslash${\tt{subsection}} or using $\backslash${\tt{label}} ($\backslash${\tt{appendices}} by itself
% starts a section numbered zero.)}


\bibliographystyle{IEEEtranN}
\bibliography{minimalPaths}

%\newpage

\begin{IEEEbiographynophoto}{Chong Di} received his Ph.D. degree in Information and Communication Engineering from Shanghai Jiao Tong University, Shanghai, China, in 2021. He is currently an assistant professor with the Shandong Artificial Intelligence Institute, Qilu University of Technology (Shandong Academy of Sciences). His research interests include reinforcement learning, learning automata and their applications. 
\end{IEEEbiographynophoto}
\begin{IEEEbiographynophoto}{Jinglin Zhang} received the Ph.D. degree in electronics and communication engineering from the National Institute of Applied Sciences, Rennes, France, in 2007, 2010, and 2013, respectively. He is currently a Professor with the School of Control Science and Engineering, Shandong University, Jinan, China. His research interests include computer vision and interdisciplinary research with pattern recognition and atmospheric science.	
\end{IEEEbiographynophoto}

\begin{IEEEbiographynophoto}{Zhenjiang Li} received his Ph.D degree in computer science and technology from Southeast University in 2017. He is an associate researcher in the Department of Radiation Physics \& Technology at Shandong First Medical University Affiliated Cancer Hospital. His expertise lies in MR imaging and MR-guided radiation therapy, with 19 recent SCI publications and 4 patents. He lead multiple national/provincial grants and have received several prestigious scientific awards in Shandong Province.
\end{IEEEbiographynophoto}

\begin{IEEEbiographynophoto}{Jean-Marie Mirebeau}
received the PhD degree from the University Pierre et Marie Curie, in 2010, prepared under the supervision of Prof. Albert Cohen, and is a former student of the Ecole Normale Superieure of Paris. He is a director of research in applied mathematics from the French CNRS (National Center for Scientific Research), working with Centre Borelli of the Ecole Normale Superieure of Paris-Saclay, University Paris-Saclay. His main subject of research is the numerical analysis of PDEs, focusing on difficulties related to strong anisotropies, which are addressed using tools from discrete geometry. He also studies PDE-based image analysis, and distributes open source numerical codes of his work. He received in 2016 the 9th Popov prize, which is an international prize awarded every three years for contributions in approximation theory.
\end{IEEEbiographynophoto}
\begin{IEEEbiographynophoto}{Da Chen} received his Ph.D degree in applied mathematics from CEREMADE, University Paris Dauphine, PSL Research University, Paris, France, in 2017. From 2017 to 2019, he worked as a post-doctoral researcher at CEREMADE, University Paris Dauphine, and also at Centre Hospitalier National d'Ophtalmologie des Quinze-Vingts, Paris, France. Now he is working at CEREMADE, Paris, France.  Now he is working at CEREMADE, CNRS, Universit\'e-PSL. His research interests include variational methods, machine learning, minimal paths, and geometric methods with applications in image analysis and robotics.
\end{IEEEbiographynophoto}
\begin{IEEEbiographynophoto}{Laurent D. Cohen} was student at the Ecole Normale Superieure, rue d'Ulm in Paris, France, from 1981 to 1985. He received the Master's and Ph.D. degrees in applied mathematics from University of Paris 6, France, in 1983 and 1986, respectively. He got the Habilitation \'a diriger des Recherches from University Paris $9$ Dauphine in 1995. From 1985 to 1987, he was member at the computer graphics and image processing group at Schlumberger Palo Alto Research, Palo Alto, California, and Schlumberger Montrouge Research, Montrouge, France, and remained consultant with Schlumberger afterward. He began working with INRIA, France, in 1988, mainly with the medical image understanding group EPIDAURE. He obtained in 1990 a position of Research Scholar (Charge then Directeur de Recherche 1st class) with the French National Center for Scientific Research (CNRS) in the Applied Mathematics and Image Processing group at CEREMADE, Universite Paris Dauphine, Paris, France. His research interests and teaching at university are applications of partial differential equations and variational methods to image processing and computer vision, such as deformable models, minimal paths, geodesic curves, surface reconstruction, image segmentation, registration and restoration. He is currently or has been editorial member of the Journal of Mathematical Imaging and Vision, Medical Image Analysis and Machine Vision and Applications. He was also member of the program committee for about 50 international conferences. He has authored about 260 publications in international Journals and conferences or book chapters, and has 6 patents. In 2002, he got the CS 2002 prize for Signal and Image Processing. In 2006, he got the Taylor \& Francis Prize: "2006 prize for Outstanding innovation in computer methods in biomechanics and biomedical engineering." He was 2009 laureate of Grand Prix EADS de l'Academie des Sciences in France. He was promoted as IEEE Fellow 2010 for contributions to computer vision technology for medical imaging.
\end{IEEEbiographynophoto}
\end{document}